\documentclass[sigconf]{acmart}
\renewcommand\footnotetextcopyrightpermission[1]{} 
\usepackage{color}
\usepackage{paralist}

\usepackage[toc,page]{appendix}

\newcommand{\todo}[1]{}
\renewcommand{\todo}[1]{{\color{red} {#1}}}

\graphicspath{{./figs/}}

\begin{document}

\keywords{Remote sensing, population density, dasymetric mapping}


\title[Weakly Supervised Spatial Density Estimation from Satellite Imagery]{A Weakly Supervised Approach for Estimating Spatial Density Functions from High-Resolution Satellite Imagery}

\author{Nathan Jacobs}
\authornote{Work was conducted while Nathan Jacobs was a Visiting Research Scientist at Orbital Insight, Inc.}
\affiliation{%
\institution{Computer Science, University of Kentucky}
}
\email{jacobs@cs.uky.edu}

\author{Adam Kraft}
\affiliation{%
\institution{Orbital Insight, Inc.}
}
\email{adam@orbitalinsight.com}

\author{Muhammad Usman Rafique}
\affiliation{%
\institution{Electrical Engineering, University of Kentucky}
}
\email{usman.rafique@uky.edu}

\author{Ranti Dev Sharma}
\affiliation{%
\institution{Orbital Insight, Inc.}
}
\email{ranti.sharma@orbitalinsight.com}

\begin{abstract}

We propose a neural network component, the regional aggregation layer,
that makes it possible to train a pixel-level density estimator using
only coarse-grained density aggregates, which reflect the number of
objects in an image region.  Our approach is simple to use and does
not require domain-specific assumptions about the nature of the
density function.  We evaluate our approach on several synthetic
datasets.  In addition, we use this approach to learn to estimate
high-resolution population and housing density from satellite imagery.
In all cases, we find that our approach results in better density
estimates than a commonly used baseline.  We also show how our housing
density estimator can be used to classify buildings as residential or
non-residential.

\end{abstract}

\maketitle

\section{Introduction}

\begin{figure}

\includegraphics[width=.95\linewidth]{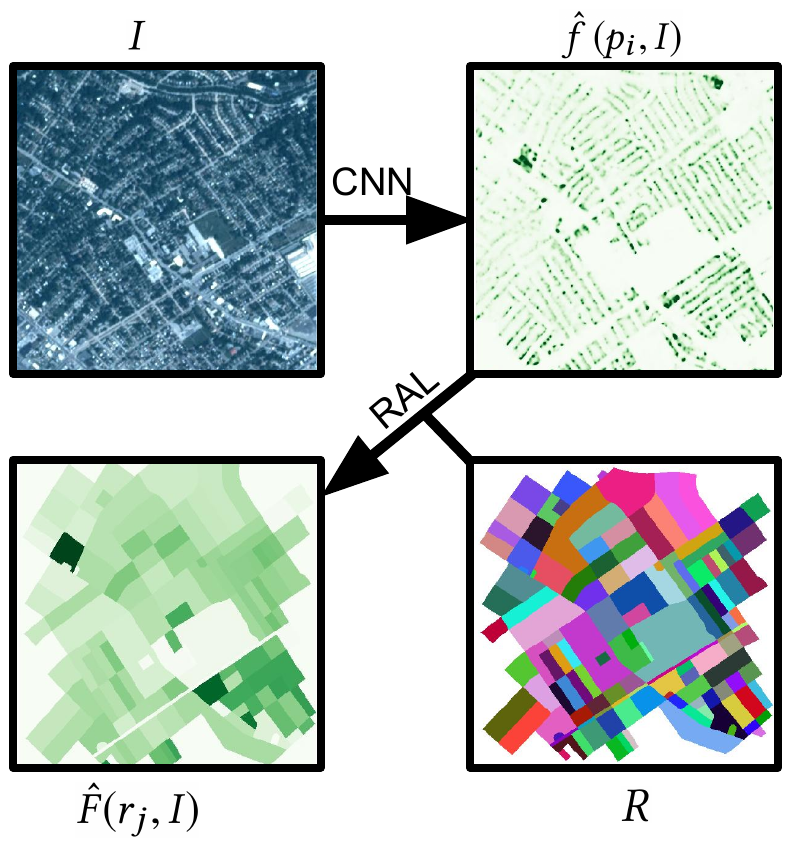}

\caption{We use a CNN to generate a pixel-level density map
  (top-right) from an input satellite image (top-left). We introduce
  the regional aggregation layer ({\em RAL}), which explicitly models
  the aggregation process for a given set of regions (bottom-right),
to generate the corresponding spatially aggregated densities
(bottom-left). This enables us to train the CNN using only aggregated
densities as ground truth.}

\label{fig:overview}
\end{figure}

The availability of high-resolution, high-cadence satellite imagery
has revolutionized our ability to photograph the world and how it
changes over time.  Recently, the use of convolutional neural networks
(CNNs) has made it possible to automatically extract a wide variety of
information from such imagery at scale. In this work, we focus on
using CNNs to make high-resolution estimates of the distribution of
human population and housing from satellite imagery.  Understanding
changes in these distributions is critical for many applications, from
urban planning to disaster response.  Several specialized systems that
estimate population/housing distributions have been developed,
including LandScan~\cite{dobson2000landscan} and the High Resolution
Settlement Layer~\cite{hrsl}. While our work addresses the same task,
our approach should be considered as a component that could be
integrated into such a system rather than a direct competitor.

Our method can be used to train a CNN to estimate any high-resolution
geospatial density function from satellite imagery.  The key challenge
in training such models is that data that reflects geospatial
densities is often coarse grained, provided as aggregates over large
spatial regions instead of at the pixel-level grid of the
available imagery.  This means that traditional methods for training
CNNs to make pixel-level predictions will not work.
To overcome this, we propose a weakly supervised learning strategy,
where we use the available coarse-grained labels to learn to make
fine-grained predictions.  With our method, we can train a CNN to
generate a high-spatial resolution output, at the pixel-level if
desired, using only labels that represent aggregates over spatial regions.
See Figure~\ref{fig:overview} for a visual overview of our approach.

Many methods for disaggregating spatial data have been developed to
address this problem, one of the most prominent is called dasymetric
mapping~\cite{wright1936method}.  The key downside of this approach is
that it requires aggregated sums at inference time. It also typically
involves significant domain-specific assumptions and intimate
knowledge of the input data and imagery.  In contrast, we propose an
end-to-end strategy for learning to predict the disaggregated
densities which makes minimal assumptions and can be applied to a wide
variety of input data. In addition, our method only requires
aggregated sums during the CNN training phase. During inference, only
a single forward pass through the CNN is required to estimate the
spatial density. It is also possible, if the aggregated sums are
available at inference time, to incorporate them using dasymetric
mapping techniques.

Our approach is easy to apply and operates quickly at training and
inference time, especially using modern GPU hardware. At training
time, there is a small amount of overhead, because we need to compute
an estimate of the aggregated densities, but at inference time no
additional computation is required.  In addition, our approach is
quite general, and could naturally be extended to area-wise averaging
and other aggregation operations, such as regional maximization. The
only requirement is that the aggregation operation must be
differentiable  w.r.t.\ the input density map.  Critically, it is not
required that the aggregation operation be differentiable w.r.t.\ any
parameters of the aggregation operation (e.g., the layout of the
spatial regions).

The main contributions of this work include: (1) proposing an end-to-end optimizable method that makes it possible to train a CNN that generates pixel-level density estimates using aggregated densities of arbitrary regions; (2) an evaluation of this method, with various extensions, including
different regularization schemes on synthetic data; (3) an evaluation
of this method on a large-scale dataset of satellite imagery for the
tasks of population and housing density estimation; (4) a demonstration of its use as a method for dasymetric mapping, which is useful if aggregated sums are available for the test data; and (5) a demonstration of how we can use our high-resolution housing density estimates, in conjunction with an existing building segmentation CNN, to perform residential vs.\ non-residential building segmentation.

\section{Related Work}

Over the past ten years, the use of deep convolutional neural networks
has rapidly advanced the state of the art on many tasks in the field
of computer vision.  For example, the top-1 accuracy for the
well-known ImageNet dataset~\cite{imagenet_ILSVRC15} (ILSVRC-2012
validation set) has risen from around $60\%$ in 2012~\cite{alexnet2012} to
$82.7\%$ in 2018~\cite{zoph2018learning}.  Significant improvements, using similar methods, have been achieved in object detection~\cite{redmon2017yolo9000}, semantic segmentation~\cite{ronneberger2015u,zhao2017pyramid,peng2017large,
deeplabv3plus2018}, and instance
segmentation~\cite{he2017mask,chen2018masklab}.  Recently, these
methods have been applied to traditional remote sensing tasks, largely
by adapting methods initially developed in the computer vision
community.  Notable examples include applications to land cover
segmentation~\cite{kussul2017landcover,pascual2018uncertainty}, land
degradation estimation~\cite{kussul2017land}, and crop
classification~\cite{ji20183d,bargiel2017new}.

\subsection{Weakly Supervised Learning}

In traditional machine learning, we are given {\em strong} labels,
which directly correspond to the desired output. In weakly supervised
learning, the provided labels are coarser grained, either semantically
or spatially, and potentially noisy. For example, Zhou et
al.~\cite{zhou2016learning} introduce a discriminative localization
method that uses image-level labels, intended for classification
training, to enable object localization and detection using a global
average pooling (GAP)~\cite{GAP_lin2013network} layer. Khoreva et al.\
present a weak supervision method for training a pixel-level semantic
segmentation CNN, using only bounding boxes~\cite{khoreva2017simple}.
Zhou et al.~\cite{zhou2018weakly} show that image-level labels can be
used for instance and semantic segmentation by finding peak responses
in convolutional feature maps. In this paper, we have supervision of
coarse spatial labels but we want to predict fine, per-pixel
prediction without access to such data for training.

The technique we propose is most closely related to the use of a GAP
layer for discriminative object localization~\cite{zhou2016learning}.
The idea in the previous work is to have a fully convolutional CNN
output per-pixel logits, average these logits across the {\em entire
image}, and then pass the averaged logits through a {\em softmax} to
estimate a distribution over the desired class labels. Once this
network is trained, it is possible to modify the network architecture
to extract the pixel-level logits and use them for object
localization. This is a straightforward process because the GAP layer
is linear.  Our proposed regional aggregation layer (RAL) is also
linear, but aggregates values from sub-regions of the image, which is
critical because of significant variability in region sizes and
shapes. Since our focus is on estimating a pixel-level density using
aggregated densities, we perform aggregation on a single channel at a
time and omit the {\em softmax}. Similarly to the previous work, after
our network is trained, we can use the intermediate pixel-level
outputs to extract the desired information at the pixel level.

\subsection{Image-Based Counting}

An important application that has received comparatively less
attention is that of image-based counting. Broadly, there are two
categories of image-based counting techniques: direct and indirect.
Direct counting methods are usually based on object detection,
semantic segmentation, or subitizing. Indirect methods are trained to
predict object densities which can be used to estimate the count in an
image. Typical object detection/segmentation based methods used for
counting have to exactly locate each instance of an object which is a
challenging task due to varying scales of objects in images. Several
methods formulate counting as a subitizing problem, inspired by
ability of humans to estimate the count of objects without explicitly
counting. Zhang et al.~\cite{zhang2015salient} proposed a
class-agnostic object-counting method, salient object subitizing,
which uses a CNN to predict image-level object counts. The work by
Chattopadhyay et al.~\cite{chattopadhyay2016counting} tackles multiple
challenges including multi-class counting in a single image and
varying scales of objects. These methods require strongly-annotated
training data. Gao et al.~\cite{gao2018c} propose a weakly supervised
framework that uses the known count of objects in images to enable
object localization. 

Density-based counting methods are generally used to tackle
applications which have a large number of objects and drastically
varying scales of object instances. For example, in crowd counting, a
person might span hundreds of pixels if they are close to the camera
or only a few if they are in the distance and occluded by others.
Regardless, both must be counted. Recent CNN-based methods have been
shown to provide state-of-the-art results on large-scale crowd
counting. Hydra CNN~\cite{onoro2016towards} is a density estimation
method based on image patches. A contextual pyramid CNN by Sindagi and
Patel~\cite{sindagi2017generating} leverages global and local context
for crowd density estimation. In the previous work, counting is
formulated so that all the density values in an image should sum up to
the count of objects. Our work is similar in that we predict
pixel-level densities, but we focus on estimating densities from
overhead views and using arbitrarily defined regions as training data.

\subsection{Dasymetric Mapping}

If aggregated sums for the area of interest are available at inference
time, then an approach known as {\em dasymetric mapping} can be used
to convert these summations to pixel-level densities. There is a long
history of using dasymetric mapping techniques to disaggregate
population
data~\cite{holt2004dasymetric,stevens2015disaggregating,gaughan2015exploring,sorichetta2015high,pavia2017can}.
These works have typically focused on applying traditional machine
learning techniques to low-resolution satellite imagery (or other
auxiliary geospatial data, such as land-cover classification).

The seminal work of Wright~\cite{wright1936method} discusses the
high-level concepts of inhabited and uninhabited areas and how to
realistically display this information on the maps. In addition, a
method of calculating the population densities based on geographical
information of populations is also presented. Usage of grid cells and
surface representations for population density visualization is
proposed by Langford and Unwin~\cite{langford1994generating}. This
work shows how a dasymetric mapping method can be used to more
accurately present density of residential housing. Fisher and
Langford~\cite{fisher1996modeling} show how areal interpolation can be
used to transform data collected from one set of zonal units to
another. This makes it possible to combine information from different
datasets. Eicher and Brewer~\cite{eicher2001dasymetric} show that
areal manipulation methods (such as the one by Fisher and
Langford~\cite{fisher1996modeling}) can be leveraged to generate
dasymetric mapping by combining the choropleth and land-use data. In
the work by Mennis~\cite{mennis2003generating}, areal weighting and
empirical sampling are proposed to estimate population density. This
work employs surface representations to make dasymetric maps based on
aerial images and US Census data. A classical machine learning-based
algorithm (random forest) is suggested by Gaughan et
al.~\cite{gaughan2015exploring} for disaggregating population
estimates. The proposed model considers the spatial resolution of the
census data for better parameterization of population density
prediction models.

Also, while we don't require population counts
for the area of interest at test time, we show in Section~\ref{sec:dasymetric} how
we can use our per-pixel density estimates to perform dasymetric
mapping.  This could potentially work better than our raw estimates if
the population counts are accurate and our density estimates are
biased. 

\subsection{Learning-Based Population Density Mapping}

There have been several works that explore the use of deep learning
for population density mapping.  These approaches have the advantage
that they can estimate population density directly from the imagery,
without requiring regional sums to disaggregate.  Doupe et
al.~\cite{doupe2016equitable} and Robinson et
al.~\cite{robinson2017deep} propose to estimate the population in a
LANDSAT image patch using the VGG~\cite{vgg} architecture. Because of
the mismatch in spatial area of the image tiles and the population
counts, they both make assumptions about the distribution of people in
a region as a pre-processing step.  Our work differs in that we
provide pixel-level population estimates, use higher-resolution
imagery, and allow the end-to-end optimization process to learn the
appropriate distribution within each region.  More similar to our
approach, Pomente and Aleandri~\cite{pomente2017} propose using a CNN
to generate pixel-level population predictions. While they initially
distribute the population uniformly in the region, they use a
multi-round training process to iteratively update the training data
to account for errors in previous rounds.

\section{Problem Statement}

For a given class of objects, we address the problem of estimating a
geospatial density function, $f(l)\in\mathcal{R}^+$, for a location,
$l\in\mathcal{R}^2$, from satellite imagery, $I$, of the location. This
function, $f$, reflects the (fractional) number of objects at a
particular location.  For convenience, we redefine this as a gridded
geospatial density function, $f(p_i)$, where each value corresponds to
the number of objects in the area imaged by a pixel, $p_i\in P$.
Therefore, the problem reduces to making pixel-level predictions of
$f$ from the input imagery.

The key challenge in training our model is that we are not given
samples from the function, $f$.  Instead, we are only given a set of
spatially aggregated values, $Y=\left\{ y_1\ldots y_n\right\}$, where
each value, $y_i \in \mathcal{R}^+$, represents the number of objects
in the corresponding region, $r_i\subset P$. Specifically,
we define $y_i = F(r_i) = \sum _{p_j\in r_i}f\left( p_j\right)$. These
regions, $R=\left\{ r_1\ldots r_n\right\}$, could have arbitrary
topology and be overlapping, but in practice will typically be simply
connected and disjoint.  Also, in many cases the labels will be
non-negative integers, but we generalize the formulation to allow them
to be non-negative real values.  This generalization does not impact
our problem definition or algorithms in any significant way.

To solve this problem, we must minimize the difference between the
aggregated labels, $F$, and our estimates of the labels,
$\hat{F}(r_{i}) = \sum_{p_j\in r_i}\hat{f}\left( p_j\right)$, where $\hat{f}$ is the estimated density function.  See
Figure~\ref{fig:overview} for a visual overview of this process.  Our
key observation is that the aggregation process is linear w.r.t.\ the
geospatial density function, $f$.  This means that we can propagate
derivatives through the operation, enabling end-to-end optimization of
a neural network that predicts $\hat{f}$.

\section{Methods}

Our goal is to estimate a per-pixel density function, $\hat{f}(p_i)$,
from an input satellite image, $I$.  We represent the model we are
trying to train as, $D(I,p_i;\Theta) = \hat{f}(p_i) \approx f(p_i)$,
where $\Theta$ is the set of all model parameters.  We propose to
implement this model as a fully convolutional neural net (CNN) that
generates an output feature map of the same dimensionality as the
input image.  We assume that the spatial extent of the image is known,
therefore we can compute the spatial extent of each pixel.  This makes
it possible to determine the degree of overlap between regions and
pixels.  For simplicity, we will assume that the input image and
output feature map have the same spatial extent, although this is not
a requirement.

The CNN can take many different forms, including adaptations of CNNs
developed for semantic
segmentation~\cite{zhao2017pyramid,peng2017large,deeplabv3plus2018}.
The only required change is setting the number of output channels to
the number of variables being disaggregated and, since these densities
are positive, including a {\em softplus}
activation~\cite{dugas2001incorporating} ($\log(exp(x) + 1 )$) on the
output layer. Like the rectified linear unit ({\em ReLU}), output of
softplus is always positive. However, {\em ReLU} suffers from
vanishing gradients for negative inputs. The derivative of {\em
softplus} is a {\em sigmoid} and it offers nonzero values for both
positive and negative inputs. As we show in
Section~\ref{sec:evaluation}, this activation can also be replaced
with a {\em sigmoid} if the output has a known upper bound as well.

\subsection{Regional Aggregation Layer (RAL)}

Since per-pixel training data is not provided, we need to find a way
to use the provided aggregated values.  Our solution is to implement
the forward process of regional aggregation directly in the deep
learning framework, as a novel differentiable layer.  The output of
this regional aggregation layer (RAL) is an estimate of the aggregated
value, $\hat{F}$, for each region in $R$.  The input to this layer is
the output of the CNN, $D(I,p_i;\Theta)$, defined in the previous
section.

To implement this layer, we first vectorize the output of the CNN,
$\mathop{vec}(\hat{f}) \in \mathcal{R}^{HW \times 1}$ for an image of
height, $H$, and width, $W$. We multiply this by a regional
aggregation matrix, $M$. Given $n$ regions, the aggregation operation
can be written as, $ \hat{F} = M \mathop{vec}(\hat{f})$,
where $\hat{F}\in \mathcal{R}^{n \times 1}$ is a vector of aggregated
values of all regions and $M \in \mathcal{R}^{n \times HW}$. Defining
$M = [M_1, M_2, \cdots, M_n]^T$, the aggregated value of a region,
$r_i$, is given as $F_i = M_i \mathop{vec}(\hat{f})$. This means that
the aggregation matrix, $M$, has a row for each spatial region and a
column for each pixel. The elements of $M$ correspond to the proportion
of a particular pixel that should be assigned to a given region.

This formulation allows for soft assignment of pixels and for overlapping regions, where pixels are assigned to multiple regions. We expect that regions will be larger
than the size of a pixel in our input image, typically much larger.
For simplicity, we have assumed that each region, $r_i$, is fully
contained inside of one image. In the next section, we describe our strategy for 
handling regions that don't satisfy this assumption.

The proposed layer involves a dense matrix multiplication. One negative aspect is that if an image includes many pixels and 
many regions the matrix is likely to consume a significant amount of
memory. In addition, since the values will be mostly zeros, much of
the computation will be wasted. Some of this could potentially be
mitigated by using a sparse matrix representation. 

\subsection{Model Fitting}

We optimize our CNN to predict the given labels, $Y$, over regions,
$R$.  In doing so, we obtain our goal of being able to predict a
pixel-level density function, $\hat{f}(p_i)$. Our loss function is
defined as follows:
\begin{equation}
    J(\Theta) = \sum_{j=1}^{N} \left\| y_j - \sum_{p_i \in r_j}
    D(\mathbf{I},p_i;\Theta) \right\|. \nonumber
\end{equation}
We use the $L_1$ norm in this work, but using others is
straightforward.  Additionally, if we know that the underlying density
function has a particular form, such as that it is sparse or
piecewise constant, we can easily add additional regularization
terms, such as $L_1$ or total variation, directly to output of our
CNN, $D$. Since all components are differentiable, we can use standard deep-learning techniques for optimization. We show examples of this
applied to synthetic and real-world data in
Section~\ref{sec:evaluation}.

One important implementation detail is how to handle regions that
extend beyond the boundary of the valid output region of the CNN.
This is important, because otherwise we will not be summing over the
full region when computing the aggregated value and will be biased toward
predicting a higher density than is actually present.  Therefore,
if a region is not fully contained in the valid output region of the
CNN, we ignore all pixels assigned to the region when computing the
loss.

\subsection{Dasymetric Mapping}
\label{sec:dasymetric}

The method described in the previous sections can be used to provide
density estimates for any location using only satellite imagery. However, if aggregated values are also available at inference time, we can use our density estimator to perform dasymetric mapping. The implementation is as follows: we calculate per-pixel density estimates, $\hat{f}\left(p_i\right)$, as well as region aggregate estimates, $\hat{F}\left(r_j\right)$. We then normalize the density in each region to sum to one. Using the available aggregated values, $F\left(r_j\right)$, we then adjust our pixel-level estimates such that values in the region sum to the provided value, $F\left(r_j\right)$.  Specifically, our estimate for a given pixel, $p_i\in r_j$, is
\begin{equation}
    \hat{f}_{dasy}\left(p_i\right) = 
    \hat{f}\left(p_i \right) \times \dfrac {F\left(r_j\right)}{\hat{F}\left(r_j\right)}. \nonumber
\end{equation}
The end result is that the summation
of the density in the output exactly matches the provided value, but
the values are regionally re-distributed based on the imagery. See
Section~\ref{sec:eval:census} for example outputs from using this
approach.

\section{Evaluation}
\label{sec:evaluation}

We evaluated the proposed system on several synthetic datasets and a large-scale real-world dataset using US Census data and high-resolution satellite imagery. We find that our approach is able to reliably recover the true function on our synthetic datasets and generate reasonable results on our real-world example.

\subsection{Implementation Details}

We implemented\footnote{The source code for replicating our synthetic
data experiments is publicly available at
\url{https://github.com/orbitalinsight/region-aggregation-public}.}
the proposed method and all examples in Keras,
training each on an NVIDIA K80 GPU.  The optimization approaches are
standard, and the specific optimization algorithms and training
protocol are defined with each example. We used standard training
parameters, without extensive parameter tuning.

\subsection{Synthetic Data (binary density)}

\begin{figure}

\def\arraystretch{.1}
\setlength{\tabcolsep}{.1em}
\begin{tabular}{cccccc}
 $I$ & $f$ & $R$ & $F$ & $\hat{f}$ ({\em unif}) & $\hat{f}$ ({\em RAL})\\
	\includegraphics[width=.16\linewidth]{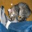} &  
    \includegraphics[width=.16\linewidth]{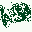} &  
    \includegraphics[width=.16\linewidth]{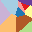} &
    \includegraphics[width=.16\linewidth]{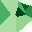} &
    \includegraphics[width=.16\linewidth]{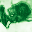} &
    \includegraphics[width=.16\linewidth]{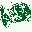} \\

	\includegraphics[width=.16\linewidth]{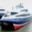} &  
    \includegraphics[width=.16\linewidth]{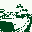} &  
    \includegraphics[width=.16\linewidth]{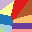} &
    \includegraphics[width=.16\linewidth]{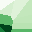} &
    \includegraphics[width=.16\linewidth]{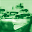} &
    \includegraphics[width=.16\linewidth]{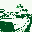} \\
    
    	\includegraphics[width=.16\linewidth]{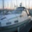} &  
    \includegraphics[width=.16\linewidth]{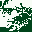} &  
    \includegraphics[width=.16\linewidth]{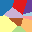} &
    \includegraphics[width=.16\linewidth]{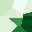} &
    \includegraphics[width=.16\linewidth]{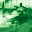} &
    \includegraphics[width=.16\linewidth]{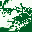} \\
    
    	\includegraphics[width=.16\linewidth]{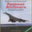} &  
    \includegraphics[width=.16\linewidth]{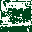} &  
    \includegraphics[width=.16\linewidth]{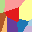} &
    \includegraphics[width=.16\linewidth]{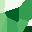} &
    \includegraphics[width=.16\linewidth]{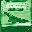} &
    \includegraphics[width=.16\linewidth]{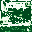} \\
    
    	\includegraphics[width=.16\linewidth]{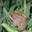} &  
    \includegraphics[width=.16\linewidth]{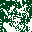} &  
    \includegraphics[width=.16\linewidth]{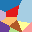} &
    \includegraphics[width=.16\linewidth]{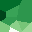} &
    \includegraphics[width=.16\linewidth]{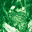} &
    \includegraphics[width=.16\linewidth]{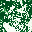} \\
    
    	\includegraphics[width=.16\linewidth]{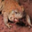} &  
    \includegraphics[width=.16\linewidth]{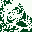} &  
    \includegraphics[width=.16\linewidth]{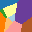} &
    \includegraphics[width=.16\linewidth]{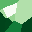} &
    \includegraphics[width=.16\linewidth]{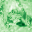} &
    \includegraphics[width=.16\linewidth]{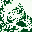} \\
    
\end{tabular}

\caption{\textbf{Synthetic disaggregation results.}  Each row shows a test example with columns representing (from left to right): the input imagery, $I$, used to generate the synthetic data, the underlying ground-truth geospatial function, $f$ (darker green is a larger value), the regions, $R$ (as unique colors), the aggregated ground-truth labels per-region, $F(r_j)$ (values filled in for each region), and the predicted geospatial functions, $\hat{f}$, using the baseline ({\em unif}) and our {\em RAL} method (using same color coding as the ground truth).}

\label{fig:synth}

\end{figure}

We first evaluated our method on a synthetic dataset constructed using the CIFAR-10 dataset~\cite{cifar10} This dataset consists of $60\,000$ $32\times 32$ images, of which $10\,000$ serve as a held-out test set.  We reserve $2\,500$ of the training examples for a validation set.  The pixel intensities were scaled between -1 and 1. To synthesize regions (Figure~\ref{fig:synth}, column 3), we randomly selected $10$ points inside the image and create the corresponding Voronoi diagram. Our method for generating a synthetic per-pixel geospatial function, $f$, is as follows: We randomly selected $15$ RGB color points from the entire CIFAR-10 dataset. For each pixel in the dataset, the value was set to one if the $L_2$-distance to any of the $15$ randomly selected points is less than a threshold ($0.2$), otherwise the value was set to zero. See Figure~\ref{fig:synth} (column 2) for example density functions.

For this example, we propose a simple CNN base architecture, which
treats each pixel independently. The CNN consists of three $1\times 1$
convolutional hidden layers (64/32/16 channels respectively, $L_2$
kernel regularization of $1\mathrm{e}{-4}$).  Each hidden layer had an
ReLU activation function and was followed by a batch normalization
layer (with a fixed scale and momentum of $0.99$). The output $1\times
1$ convolutional layer had a single channel and the {\em softplus}
activation function, to ensure non-negatively of the output (similar
results were obtained using a ReLU activation function).

We trained two such networks, one which used our regional aggregation layer to compute sums based on the provided regions ({\em RAL}) and a baseline which uniformly distributed the sums across the corresponding region ({\em unif}). We optimized both networks using the AMSGrad~\cite{reddi2018convergence} optimizer, a variant of the Adam~\cite{adam} optimizer, with an initial learning rate of $1\mathrm{e}{-2}$ and batch size of $64$. We trained for $120\,000$ iterations, dropping the learning rate by $0.5$ every $40\,000$ iterations.

Figure~\ref{fig:synth} shows example results on the test set, which
demonstrate that using our regional aggregation layer results in
significantly better estimates of the true geospatial function.
Specifically, we note that the {\em RAL} method yields per-pixel
predictions that are qualitatively more similar to the ground truth.
Quantitatively, our approach gives a mean absolute error (MAE) of
0.040 while the baseline, {\em unif}, gives an MAE of 0.39. Both are
superior to a randomly initialized network, which gives an MAE of
0.484 w/ standard deviation of 0.003 (across thirty different random
initializations).

\begin{figure}
  \includegraphics[width=\linewidth]{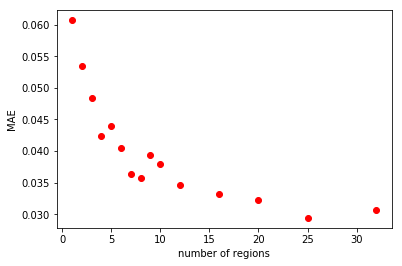}
  \caption{Increasing the number of regions per image decreases the mean absolute error (MAE) of the learned density estimator.}
  \label{fig:maevsregion}
\end{figure}

We evaluated the impact of changing the number of regions per image on the MAE of the resulting density estimates.  All other aspects are the same as the previous experiment.  Figure~\ref{fig:maevsregion} shows the results, which indicate that using a single region per image gives the highest MAE. Increasing the number of regions reduces the error, with diminishing returns after about 15 regions.  This highlights an interesting trade-off because increasing the number of regions increases the required effort in collecting training data.

\subsection{Synthetic Data (real and integer density)}

\begin{figure}

\def\tblsubwidtha{.135\linewidth}

\def\arraystretch{.1}
\setlength{\tabcolsep}{.1em}
\begin{tabular}{c|ccc|ccc|}
 $I$ & $f (count)$ & $\hat{f}$ ({\em unif}) & $\hat{f}$ ({\em RAL}) & $f (ratio)$ & $\hat{f}$ ({\em unif}) & $\hat{f}$ ({\em RAL})\\
	\includegraphics[width=\tblsubwidtha]{cifar/cifar_image_0.png} &  
    \includegraphics[width=\tblsubwidtha]{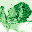} &  
    \includegraphics[width=\tblsubwidtha]{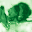} &
    \includegraphics[width=\tblsubwidtha]{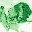} &
    \includegraphics[width=\tblsubwidtha]{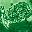} &
    \includegraphics[width=\tblsubwidtha]{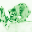} &
    \includegraphics[width=\tblsubwidtha]{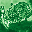} \\
    
    \includegraphics[width=\tblsubwidtha]{cifar/cifar_image_1.png} &  
    \includegraphics[width=\tblsubwidtha]{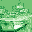} &  
    \includegraphics[width=\tblsubwidtha]{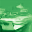} &
    \includegraphics[width=\tblsubwidtha]{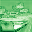} &
    \includegraphics[width=\tblsubwidtha]{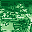} &
    \includegraphics[width=\tblsubwidtha]{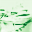} &
    \includegraphics[width=\tblsubwidtha]{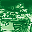} \\
    
    \includegraphics[width=\tblsubwidtha]{cifar/cifar_image_2.png} &  
    \includegraphics[width=\tblsubwidtha]{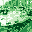} &  
    \includegraphics[width=\tblsubwidtha]{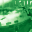} &
    \includegraphics[width=\tblsubwidtha]{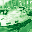} &
    \includegraphics[width=\tblsubwidtha]{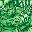} &
    \includegraphics[width=\tblsubwidtha]{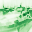} &
    \includegraphics[width=\tblsubwidtha]{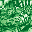} \\
    
\end{tabular}

\caption{\textbf{Results for more synthetic functions. For images $I$, we demonstrate our method on two different ground-truth geospatial functions. (column 2) Shows the ground truth for an integer counting function, $f (count)$, with the learned results in columns 3--4. (column 5) Shows the ground truth for a real valued ratio function, $f (ratio)$, with the learned  results in columns 6--7. 
}}

\label{fig:more_synth}

\end{figure}

Using the same methodology, we show that more complex functions can be learned. First, we attempt to learn a function,$ f (count)$, with a range of integers as output. Similar to before, we randomly chose colors from the entire CIFAR dataset but this time we selected $20$ random points. For each pixel, we count the total number of the random points that are within a fixed distance ($0.4$) in RGB space. The output is in the range $\left[ 0,20\right]$. Our approach, {\em RAL}, achieves an MAE of $0.267$, compared to an MAE of $0.785$ for the baseline, {\em unif}. Sample results are shown in Figure~\ref{fig:more_synth} (columns 2--4).

We also tested our method on a real-valued function. We used the same $20$ random points as before and calculated the distances to all points in the dataset. For each pixel, we sorted the $20$ distances and calculated the fraction of the minimum distance divided by the second shortest distance as our density function, $f (ratio)$. For this synthetic function, the MAEs are $0.038$ and $0.176$ for {\em RAL} and {\em unif} respectively. See Figure~\ref{fig:more_synth} (columns 5--7) for example results.

\subsection{Synthetic Data with Priors}

In the previous experiments, we only made the assumption that the geospatial density function, $f$, was non-negative, hence the use of the {\em softplus} activation on the output layer.
In some applications, we may know additional information about the form of the density, such as that it has an upper bound or that it is sparse. In this section, we investigate whether incorporating these priors into the network architecture is beneficial.

\begin{figure}
\def\tblsubwidtha{.16\linewidth}

\def\arraystretch{.1}
\setlength{\tabcolsep}{.1em}
\begin{tabular}{ccccccc}
 $I$ & $f (sparse)$ & $\hat{f}$ $(sp)$ & $\hat{f}$ $(sp$, $L_1)$ & $\hat{f}$ $(sig)$ & $\hat{f}$ $(sig$, $L_1)$\\
	\includegraphics[width=\tblsubwidtha]{cifar/cifar_image_0.png} &  
    \includegraphics[width=\tblsubwidtha]{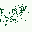} &  
    \includegraphics[width=\tblsubwidtha]{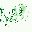} &
    \includegraphics[width=\tblsubwidtha]{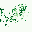} &
    \includegraphics[width=\tblsubwidtha]{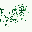} &
    \includegraphics[width=\tblsubwidtha]{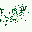} \\
    \includegraphics[width=\tblsubwidtha]{cifar/cifar_image_1.png} &  
    \includegraphics[width=\tblsubwidtha]{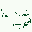} &  
    \includegraphics[width=\tblsubwidtha]{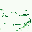} &
    \includegraphics[width=\tblsubwidtha]{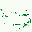} &
    \includegraphics[width=\tblsubwidtha]{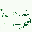} &
    \includegraphics[width=\tblsubwidtha]{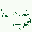} \\
    \includegraphics[width=\tblsubwidtha]{cifar/cifar_image_2.png} &  
    \includegraphics[width=\tblsubwidtha]{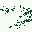} &  
    \includegraphics[width=\tblsubwidtha]{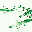} &
    \includegraphics[width=\tblsubwidtha]{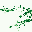} &
    \includegraphics[width=\tblsubwidtha]{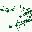} &
    \includegraphics[width=\tblsubwidtha]{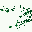} \\
    
\end{tabular}

\caption{\textbf{Enforcing priors over sparse function improves our estimate of the geospatial distribution function.} Here we use a sparse, binary ground-truth geospatial function, $f (sparse)$. Columns 3--6 show the estimated densities for models trained with the following settings: {\em softplus} activation $\hat{f}$ $(sp)$, {\em softplus} activation and $L_1$ activation regularization $\hat{f}$ $(sp$, $L_1)$, {\em sigmoid} activation $\hat{f}$ $(sig)$, {\em sigmoid} activation and $L_1$ activation regularization $\hat{f}$ $(sig$, $L_1)$. Observe that the results get progressively better as we incorporate priors that match the ground-truth function.}

\label{fig:synth_sparse}

\end{figure}

We define a new synthetic geospatial density function $f$. Our goal is for this function to be sparse and for the output range to be known (once again we use $\left[ 0,1\right]$). To create this function, we bin the  color space into 16 bins along each of the 3 dimensions of RGB, resulting in $16^{3}=4096$ total bins. We count the number of unique CIFAR images each color bin appears in, and we also calculate the average number of pixels for each time a color bin appears in an image. We select the color bins that appear in a high number of images (at least $29\,000$), but only appear in, on average, 10 pixels or less per image. This results in 26 color bins. We use these bins to create the sparse density function, $f (sparse)$, where we set pixels in the selected color bins to one, while the remaining pixels are set to zero. See  Figure~\ref{fig:synth_sparse} (column 2) for examples of this function. We use the same region generation method as before.

For training, we test two different priors. The first is the effect of using {\em sigmoid} vs.\ {\em softplus} activation. Using a sigmoid will force the outputs of our predicted density, $\hat{f}$, to be between $\left[ 0,1\right]$. The second prior is an $L_1$ penalty ($\lambda=1\mathrm{e}{-4}$) on the activations of the CNN before the {\em RAL}. The goal of this penalty is to encourage sparsity for $\hat{f}$.

We trained 4 combinations with and without using the two different priors: {\em softplus} with no activation penalty, {\em softplus} with $L_1$ activation penalty, {\em sigmoid} with no activation penalty, and {\em sigmoid} with $L_1$ activation penalty. The resulting MAEs were $0.052$, $0.035$, $0.019$, and $0.012$ respectively. For this experiment, enforcing the activation output between $\left[ 0,1\right]$ with a {\em sigmoid} performed better than the {\em softplus} activation. Using an $L_1$ activation penalty also improved results for a given activation function. This shows that  modifying the network architecture to incorporate priors on the geospatial density function, $f$, is an effective strategy for improving performance of the learned estimator.  This is a promising result; this strategy could be especially useful in real-world scenarios when training data is limited.

\subsection{Census Data Example: Population and Housing Count}
\label{sec:eval:census}

We used our proposed approach, {\em RAL}, for the task of
high-resolution population density estimation. This task could be
useful for a wide variety of applications, including improving estimates
of population between the official decennial census, by accounting for
land use changes, or estimating the population in regions without a
formal census.  We compare our method to a baseline approach, {\em
unif}, that assumes the distribution of densities within a region is
uniform.  With the {\em unif} approach, we minimize a per-pixel loss
and, therefore, do not need the region definitions during the
optimization process. All other aspects of the model, dataset, and
optimization strategy are the same.

\paragraph{Evaluation Dataset}

We built a large-scale evaluation dataset with diverse geographic
locations.  For our aggregated training labels, $F$, we used
block-group population and housing counts provided from the 2010 US
Census. We used RGB imagery from the Planet ``Dove'' satellite
constellation with a ground sample distance (GSD) of $3m$. Training
was performed using images from 11 cities across the following 
geographic subdivisions: Pacific West, Mountain
West, West North Central, West South Central, South Atlantic, and
Middle Atlantic. The total area of the training set is approximately
$14\,000 km^2$.  Our validation consists of held out tiles from these
cities, with a total area of approximately $650 km^2$.  Testing was
done on tiles from Dallas and Baltimore, with a total area of
approximately $3\,000km^2$. There are, on average, $~50$ regions per
tile and each region has, on average, an area of $20\,000 m^2$. 

\paragraph{Mini-Batch Construction}

To construct a mini-batch, we randomly sample six tiles, each is $588\times588$.  For data augmentation, we randomly flip left/right and up/down.  To normalize the images, we subtract a channel-wise mean value. We use the census block groups to define a region mask, assigning each a unique integer.  If there are more than 100 block groups for a given tile, which can happen in dense urban areas, we randomly sample 100 and assign the remainder to the {\em background} class. As ground-truth labels we use the population and housing counts for the corresponding census block groups.

\paragraph{Model Architecture} For all experiments, we use the same
base neural network architecture, which uses a {\em U-net}
architecture~\cite{ronneberger2015u}, although any pixel-level
segmentation network could be used instead.  See
Table~\ref{tbl:architecture} for details of the layers of this
architecture, except the final output layer, which varies depending on
the task. The {\em concat} layer is a channel-wise concatenation. The
{\em up}() operation is a $2\times$ up-sampling using nearest neighbor
interpolation and the {\em maxpool}() operation represents a $2\times2$
max pooling.  All convolutional layers use a $3\times3$
kernel and each, except the final one, is followed by a batch
normalization layer.  This network was pre-trained, using standard
techniques, for pixel-level classification for several output classes,
including roads and buildings.

\begin{table}[h]
\caption{The neural network architecture we use for the census data experiments.}
\begin{tabular}{|l|l|l|}
\hline
type (name) & inputs & output channels \\  
\hline
conv2d (conv1.1) & image & 32 \\
conv2d (conv1.2) & conv1.1 & 32 \\
conv2d (conv2.1) & {\em maxpool}(conv1.2) & 64 \\
conv2d (conv2.2) & conv2.1 & 64 \\
conv2d (conv3.1) & {\em maxpool}(conv2.2) & 128 \\
conv2d (conv3.2) & conv3.1 & 128 \\
conv2d (conv4.1) & {\em maxpool}(conv3.2) & 256 \\
conv2d (conv4.2) & conv4.1 & 256 \\
concat (up5)& {\em up}(conv4.2), conv3.2 & 384 \\
conv2d (conv5.1) & up5 & 128 \\
conv2d (conv5.2) & conv5.1 & 128 \\
concat (up6)& {\em up}(conv5.2), conv2.2 & 192 \\
conv2d (conv6.1) & up6 & 64 \\
conv2d (conv6.2) & conv6.1 & 64 \\
concat (up7)& {\em up}(conv6.2), conv1.2 & 96  \\
conv2d (conv7.1) & up7 & 32 \\
conv2d (conv7.2) & conv7.1 & 32\\ 
\hline
\end{tabular}
\label{tbl:architecture}
\end{table}

\begin{figure*}

\def\tblsubwidth{.197\linewidth}

\def\arraystretch{.1}
\setlength{\tabcolsep}{.1em}
\begin{tabular}{ccccc}

    & $R$ & $\hat{f}_{pop}$ (\em {RAL}) & $\hat{f}_{house}$ ({\em RAL}) & bldg.\ class \\

	\includegraphics[width=\tblsubwidth]{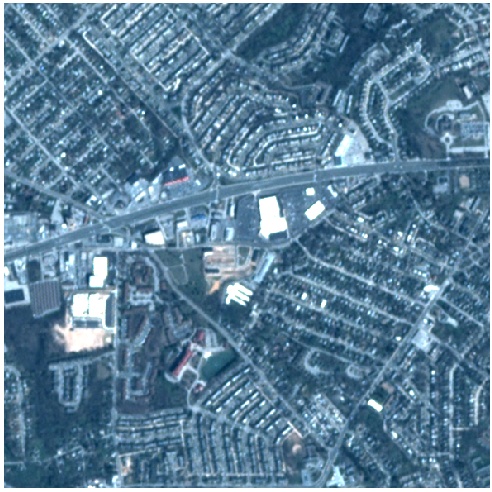} &
    \includegraphics[width=\tblsubwidth]{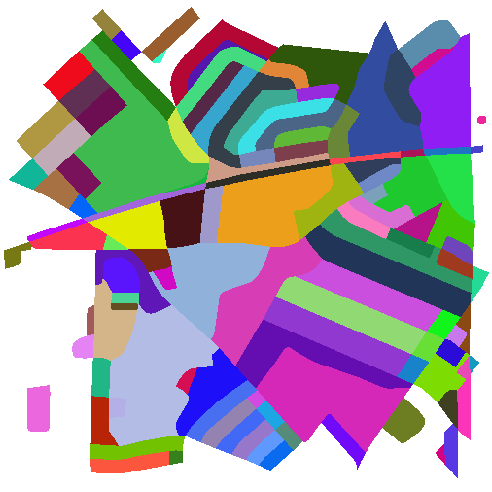} &
    \includegraphics[width=\tblsubwidth]{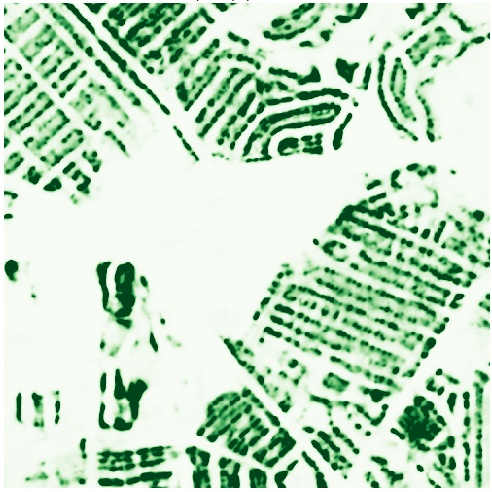} &
    \includegraphics[width=\tblsubwidth]{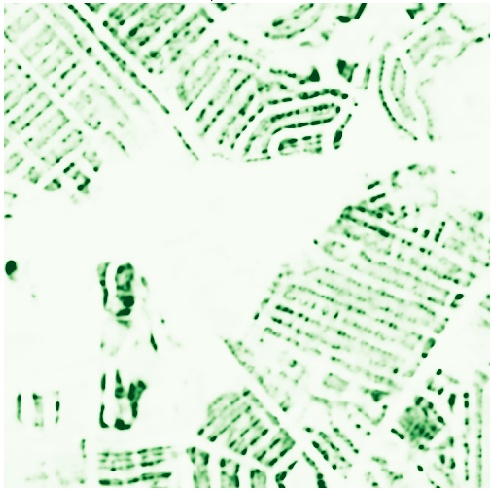} &
    \includegraphics[width=\tblsubwidth]{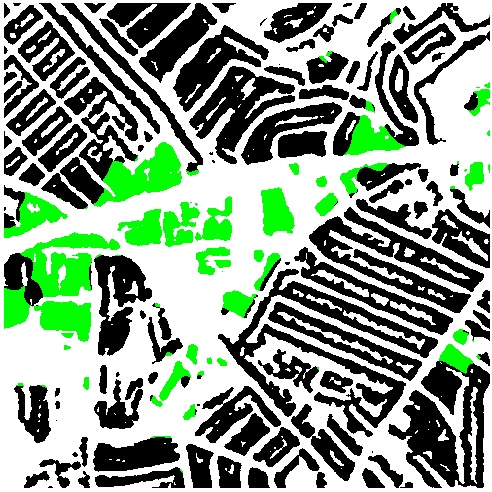} \\

	\includegraphics[width=\tblsubwidth]{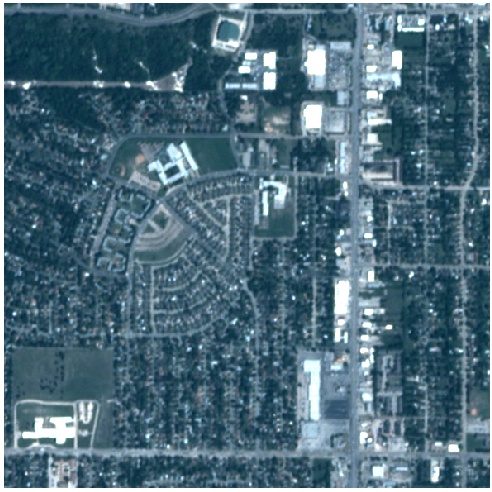} &
    \includegraphics[width=\tblsubwidth]{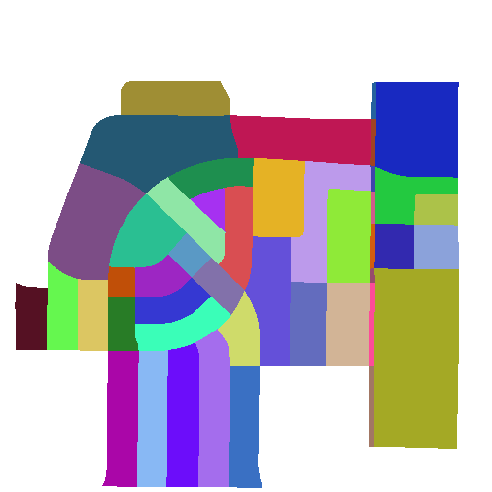} &
    \includegraphics[width=\tblsubwidth]{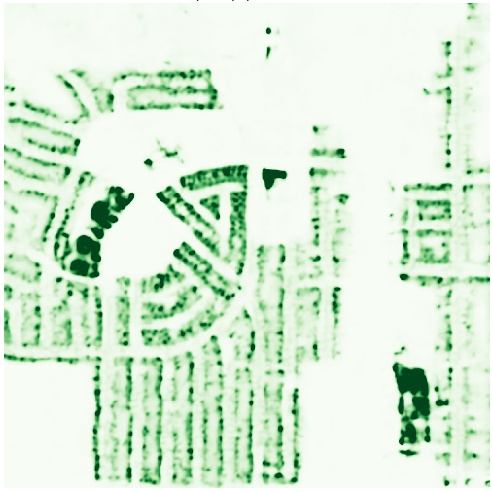} &
    \includegraphics[width=\tblsubwidth]{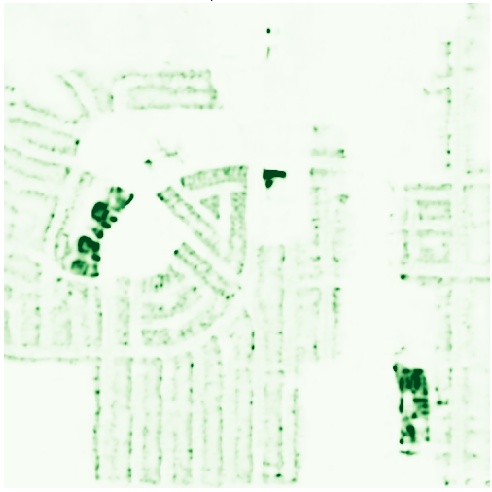} &
    \includegraphics[width=\tblsubwidth]{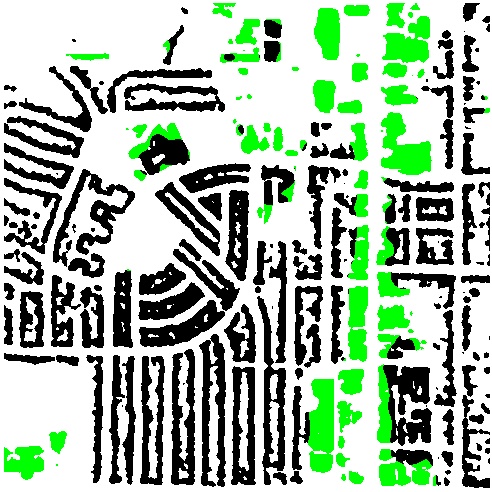} \\

	\includegraphics[width=\tblsubwidth]{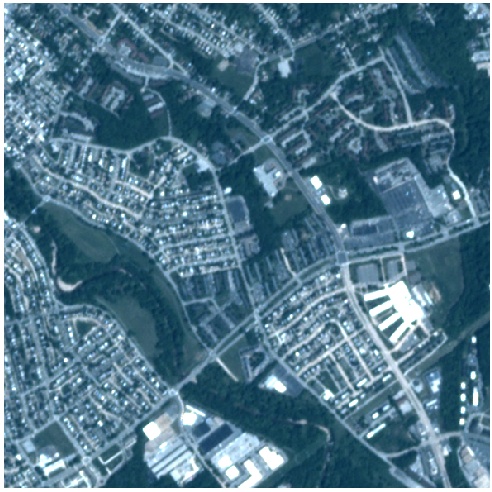} &
    \includegraphics[width=\tblsubwidth]{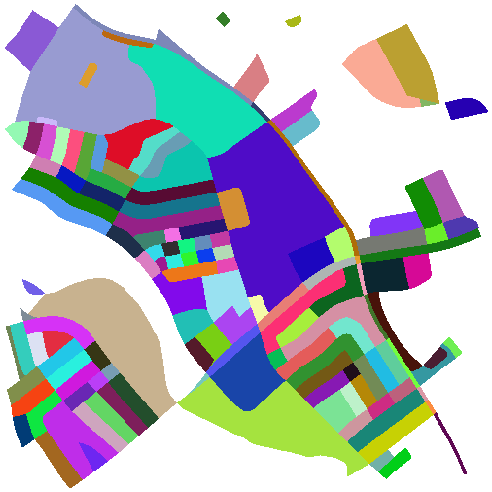} &
    \includegraphics[width=\tblsubwidth]{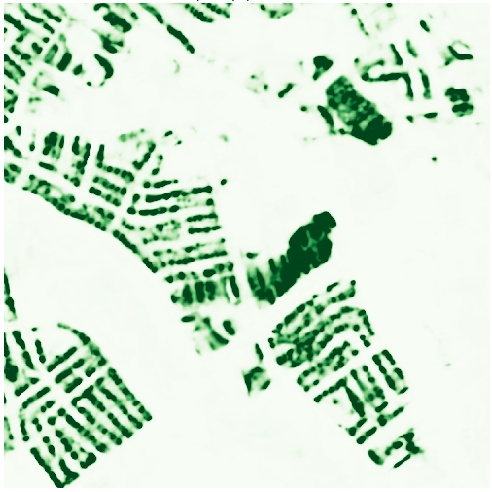} &
    \includegraphics[width=\tblsubwidth]{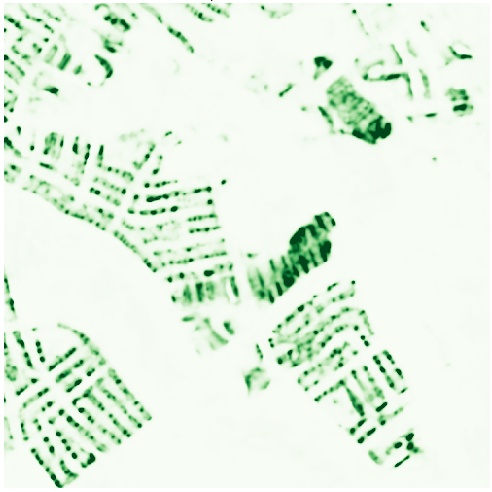} &
    \includegraphics[width=\tblsubwidth]{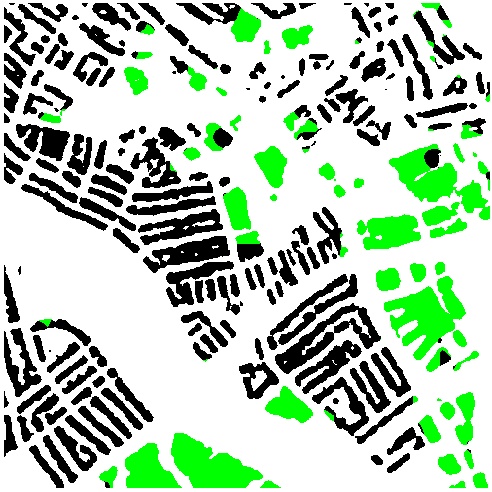} \\

	\includegraphics[width=\tblsubwidth]{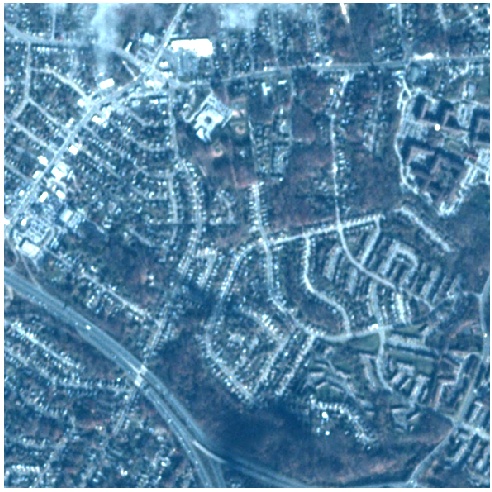} &
    \includegraphics[width=\tblsubwidth]{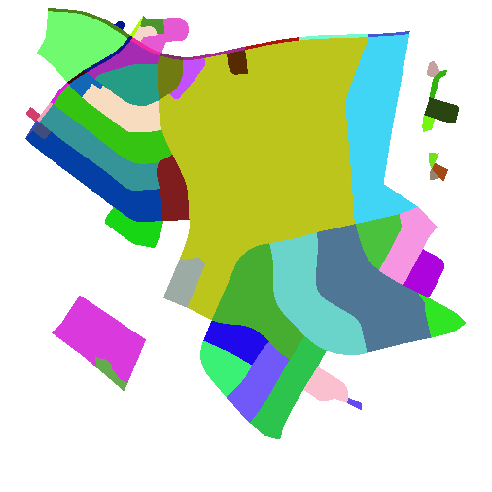} &
    \includegraphics[width=\tblsubwidth]{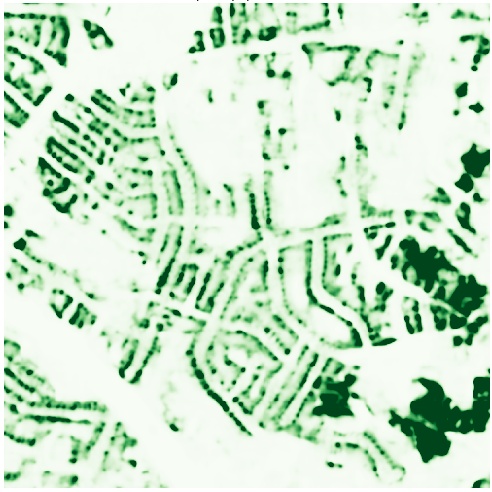} &
    \includegraphics[width=\tblsubwidth]{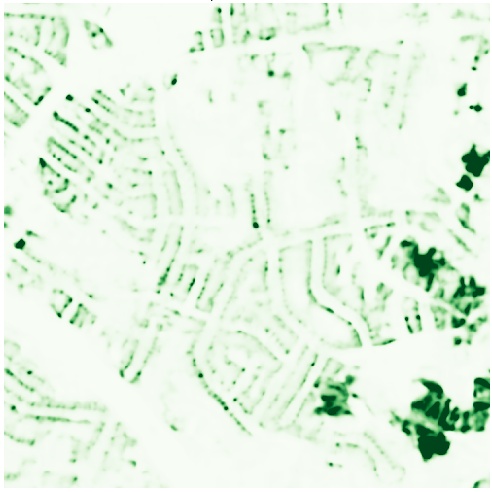} &
    \includegraphics[width=\tblsubwidth]{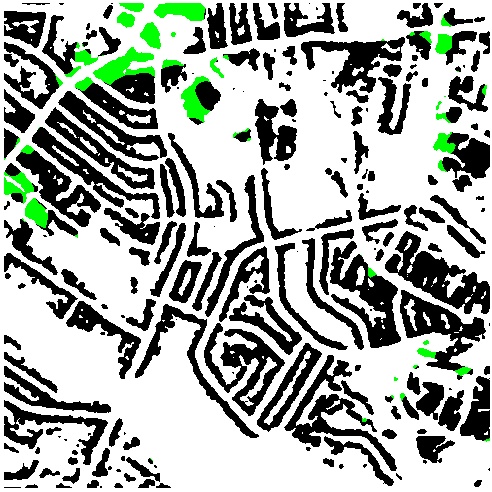} \\

	\includegraphics[width=\tblsubwidth]{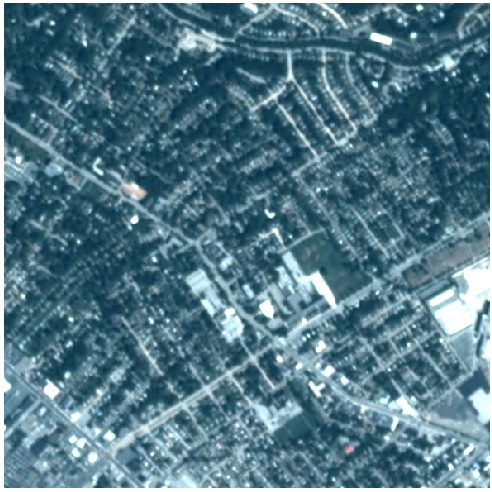} &
    \includegraphics[width=\tblsubwidth]{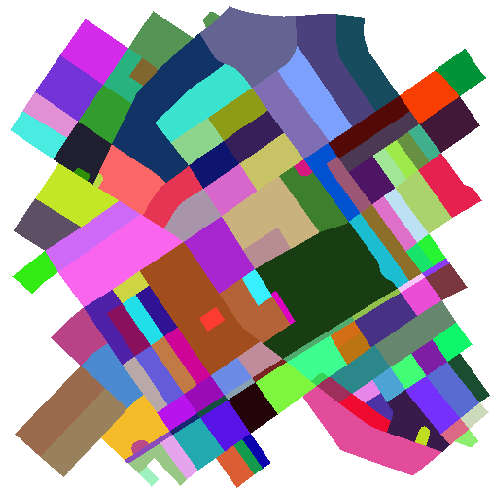} &
    \includegraphics[width=\tblsubwidth]{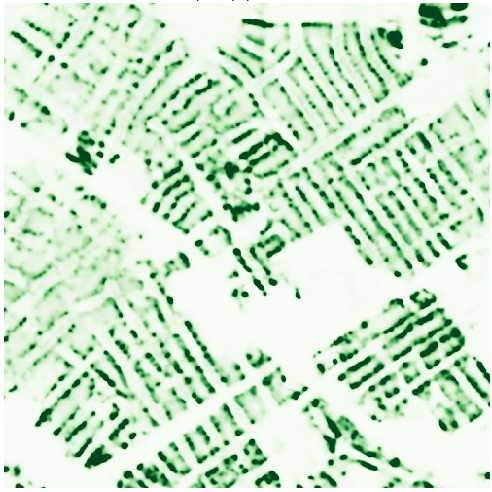} &
    \includegraphics[width=\tblsubwidth]{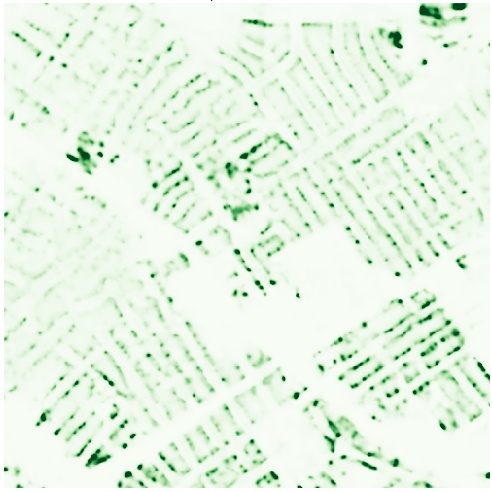} &
    \includegraphics[width=\tblsubwidth]{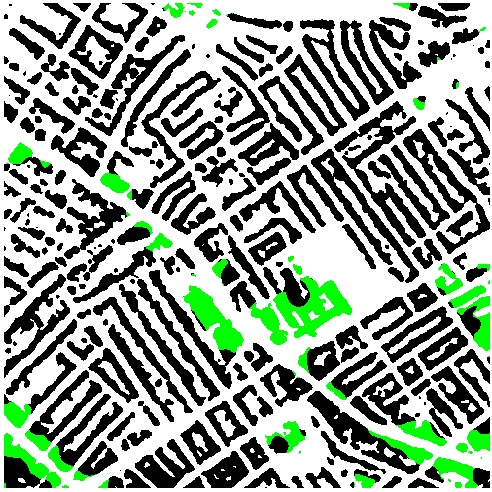} \\

\end{tabular}

\caption{Visualizations of various outputs of our model for different
    scenes. From left to right: the input image; the census block
    groups (white pixels correspond to regions that are not fully
    contained in the image); raw per-pixel density estimates for
    population and housing; and our building type classification.}

\label{fig:censusmaps}

\end{figure*}

We add two $1\times1$ convolution layers to the base network to represent the per-pixel densities, one for population and one for housing count. Each takes as input the logits of the base network and has a {\em softplus} activation function.  During training, the output of these final layers are passed to a regional aggregation layer ({\em RAL}) to generate the aggregated sums.  During inference, we remove the {\em RAL} and our network outputs the per-pixel densities for both object classes.

\paragraph{Model Fitting} We trained our model using the
AMSGrad~\cite{reddi2018convergence} optimizer, a variant of the
Adam~\cite{adam} optimizer, with a weight decay of $5\mathrm{e}{-4}$.
The initial learning rate was set to $0.001$ and we reduced the
learning rate by a factor of $0.5$ every $2\,000$ iterations. We
trained for $20\,000$ iterations, but only keep the best checkpoint,
based on validation accuracy.

\paragraph{Results} We first show qualitative output from our model.  Figure~\ref{fig:censusmaps} shows the population and housing density estimates for a satellite image. We also show the use of these estimates to do rough disambiguation between residential and non-residential buildings. To accomplish this, we first applied a pre-trained building segmentation CNN and thresholded it at $0.2$. We applied a Gaussian blur ($\sigma=4$) to our estimated house density and thresholded at $0.001$.  We then constructed a false color image where white pixels are background, black pixels correspond to buildings that we detected with a house density above the threshold (residential), and green pixels correspond to buildings with a house density below the threshold (non-residential).

\begin{figure}

\def\tblsubwidth{.327\linewidth}

\def\arraystretch{.1}
\setlength{\tabcolsep}{.1em}
\begin{tabular}{ccc}

    &$\hat{f}_{pop}$ (unif) & $\hat{f}_{pop}$ ({\em RAL}) \\

	\includegraphics[width=\tblsubwidth]{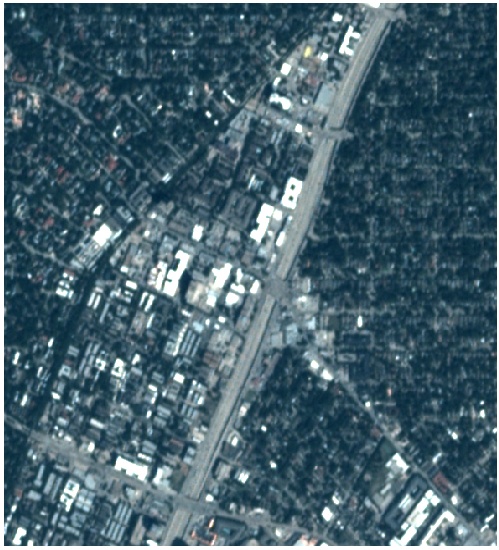} &
    \includegraphics[width=\tblsubwidth]{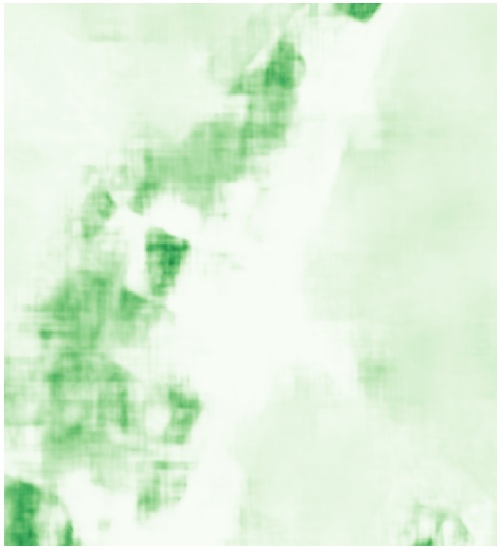} &
    \includegraphics[width=\tblsubwidth]{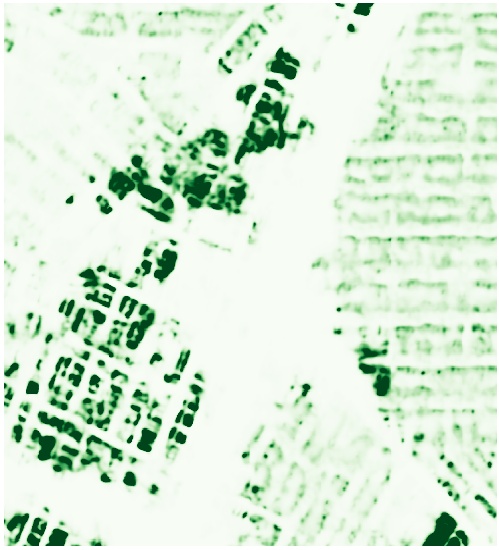} \\

	\includegraphics[width=\tblsubwidth]{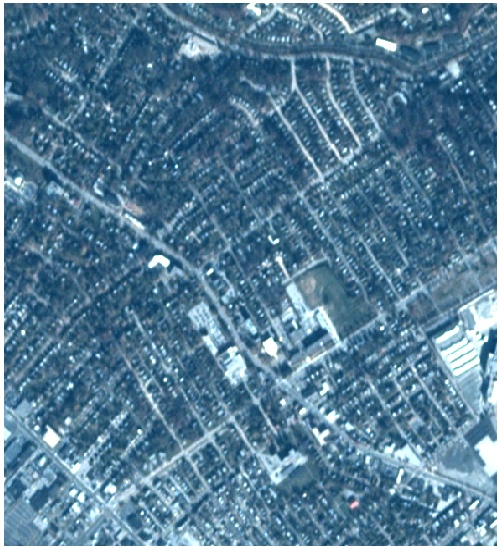} &
    \includegraphics[width=\tblsubwidth]{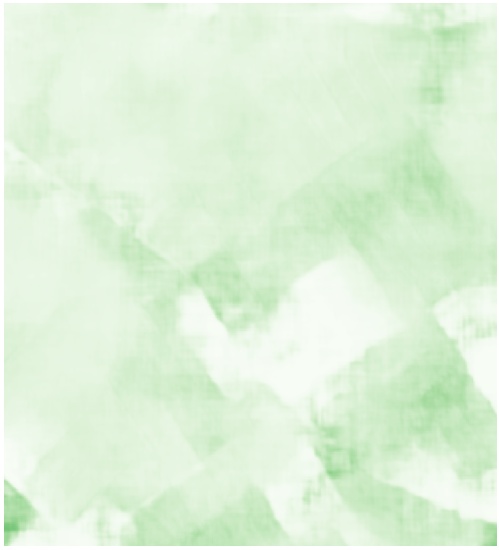} &
    \includegraphics[width=\tblsubwidth]{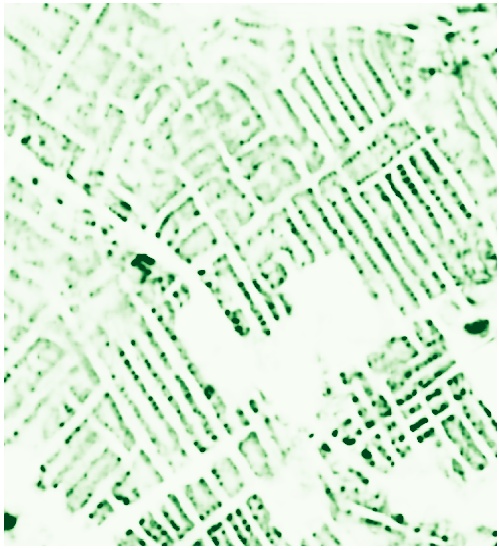} \\

	\includegraphics[width=\tblsubwidth]{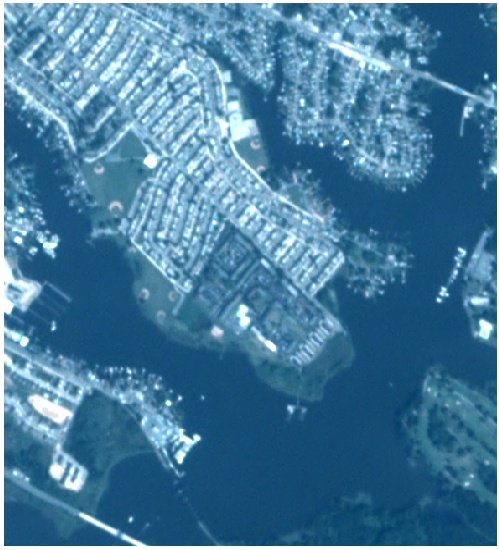} &
    \includegraphics[width=\tblsubwidth]{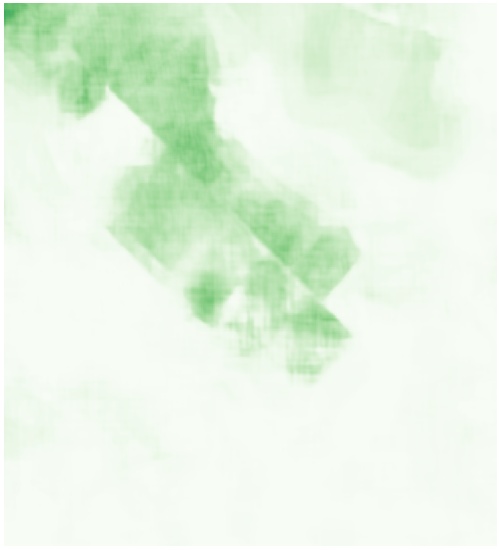} &
    \includegraphics[width=\tblsubwidth]{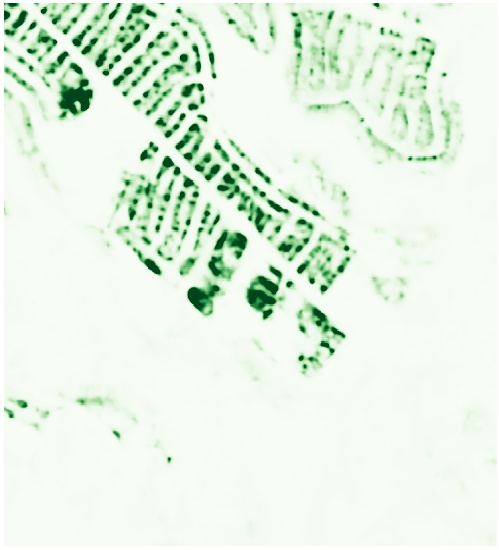} \\

	\includegraphics[width=\tblsubwidth]{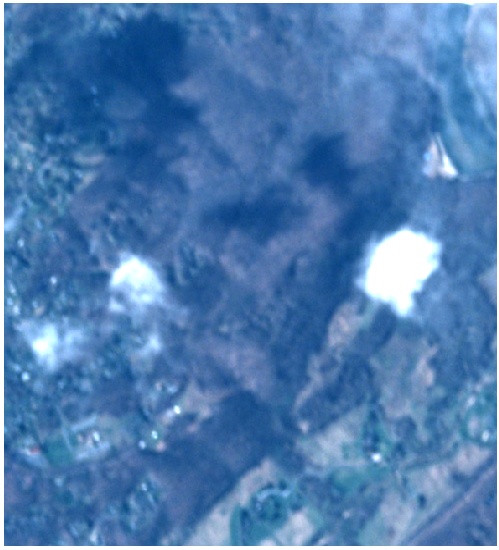} &
    \includegraphics[width=\tblsubwidth]{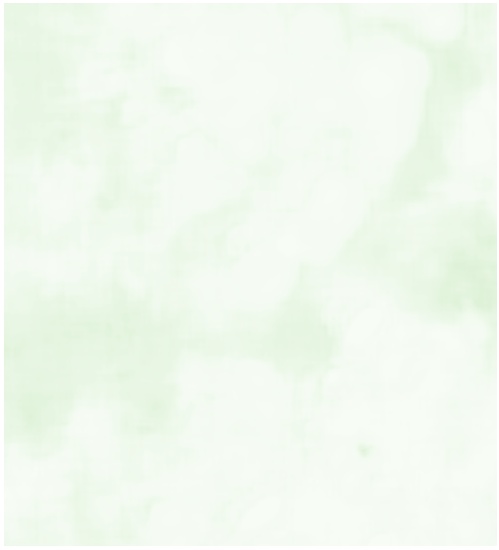} &
    \includegraphics[width=\tblsubwidth]{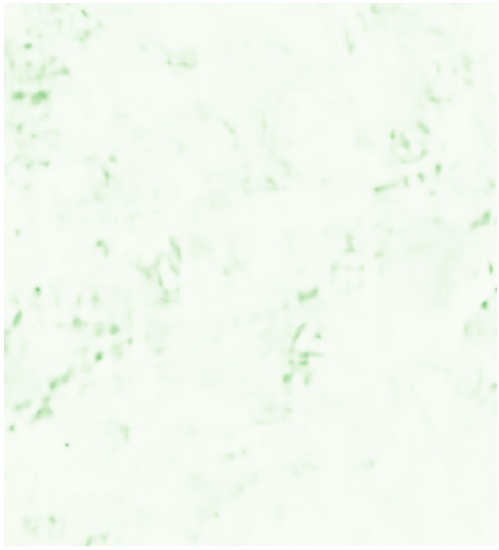} \\

	\includegraphics[width=\tblsubwidth]{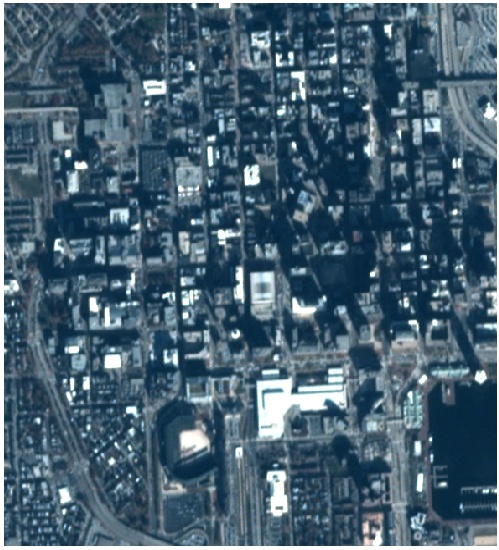} &
    \includegraphics[width=\tblsubwidth]{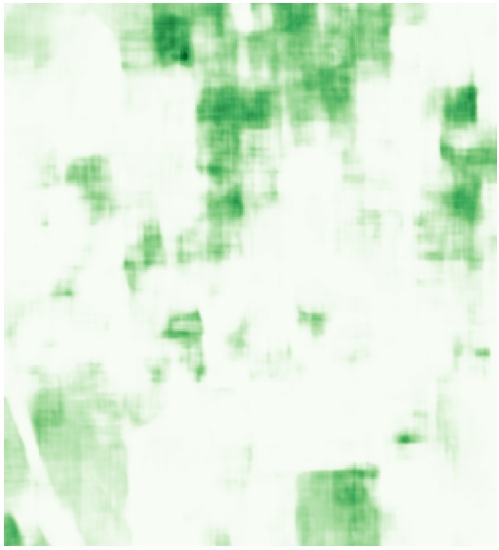} &
    \includegraphics[width=\tblsubwidth]{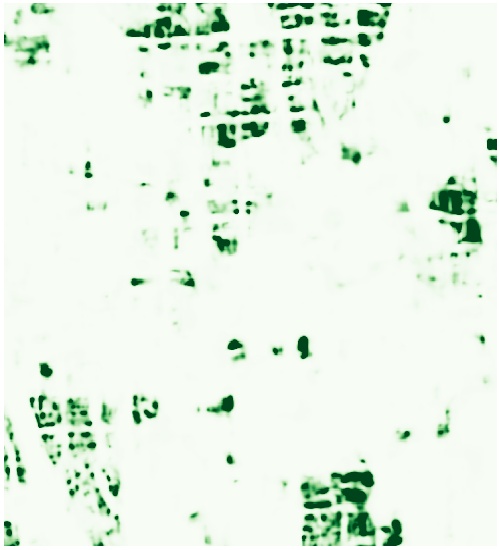} \\
\end{tabular}

\caption{A comparison of two training methods for the task of population density estimation: uniformly distributing
the aggregated sum (middle) and our method (right).}

\label{fig:censusmapsunif}

\end{figure}

For quantitative evaluation, since we don't know the true per-pixel
densities, we evaluate using the aggregated estimates on the
held-out cities. We use the pixel-level outputs from our models and
accumulate the densities for each region independently.  We find that
the uniform baseline method, {\em unif}, has an MAE of 27.7 for
population count and 12.1 for housing density. In comparison, our
method, {\em RAL}, has MAEs of 26.5 and 11.7 respectively. While these
methods are fairly close in terms of MAE, the pixel-level densities estimated by both training strategies are very different.
Figure~\ref{fig:censusmapsunif} compares these density maps.  This
shows that our approach more clearly delineates the locations of
residential dwellings, which makes this potentially more useful for
integration with different applications.

\begin{figure*}

\def\tblsubwidtha{.163\linewidth}

\def\arraystretch{.1}
\setlength{\tabcolsep}{.1em}
\begin{tabular}{ccccccc}
    & $R$ & $\hat{f}_{pop}$ ({\em RAL}) & $\hat{f}_{pop}$ (dasy) &
    $\hat{f}_{house}$ ({\em RAL}) & $\hat{f}_{house}$ (dasy) \\

	\includegraphics[width=\tblsubwidtha]{census/image_005.jpg} &
    \includegraphics[width=\tblsubwidtha]{census/region_005.png} &
    \includegraphics[width=\tblsubwidtha]{census/pop_raw_005.jpg} &
    \includegraphics[width=\tblsubwidtha]{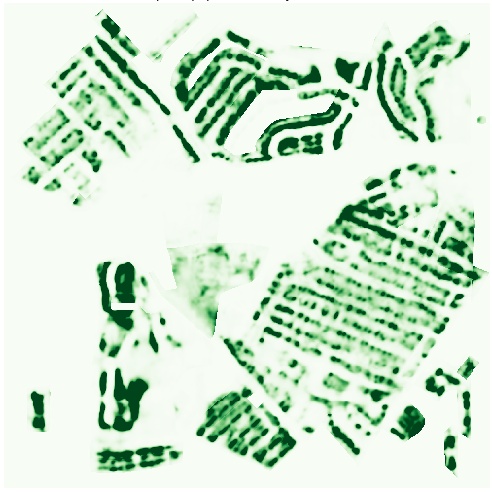} &
    \includegraphics[width=\tblsubwidtha]{census/house_raw_005.jpg} &
    \includegraphics[width=\tblsubwidtha]{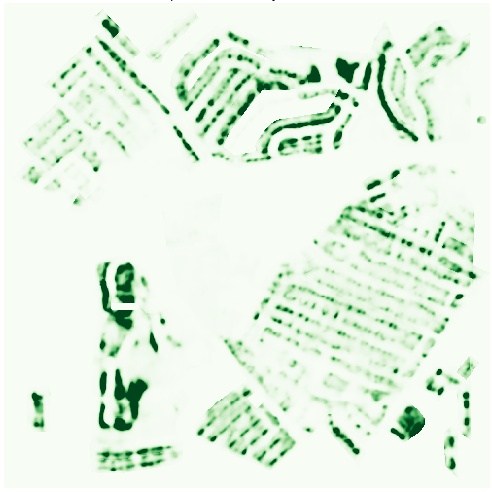} \\

	\includegraphics[width=\tblsubwidtha]{census/image_014.jpg} &
    \includegraphics[width=\tblsubwidtha]{census/region_014.png} &
    \includegraphics[width=\tblsubwidtha]{census/pop_raw_014.jpg} &
    \includegraphics[width=\tblsubwidtha]{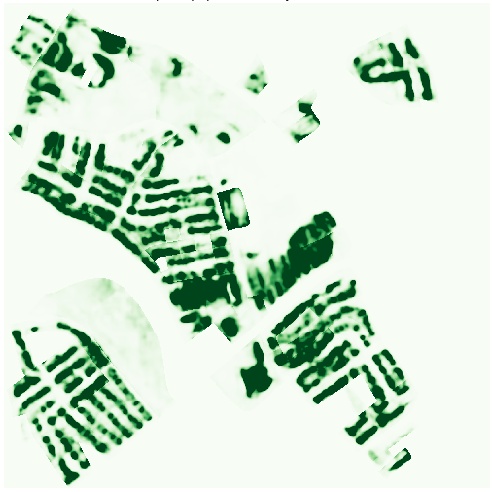} &
    \includegraphics[width=\tblsubwidtha]{census/house_raw_014.jpg} &
    \includegraphics[width=\tblsubwidtha]{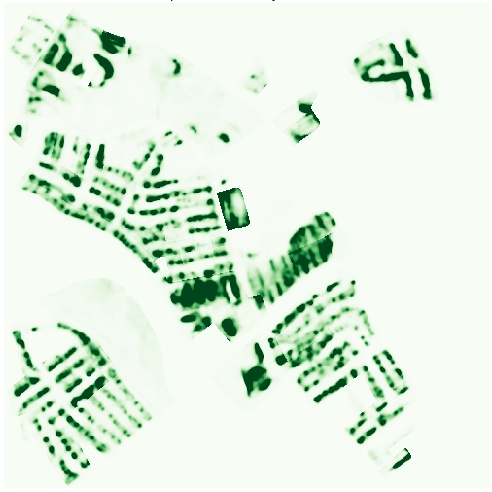} \\

\end{tabular}

\caption{Examples of using our model as a distribution estimate for dasymetric mapping. From left to right: the input image; the census block
    groups (white pixels correspond to regions that are not fully
    contained in the image); per-pixel density estimates for
    population and housing, both raw and dasymetric (note that the
    dasymetric mapping sets population densities to zero for regions
    outside the image).}

\label{fig:censusdasymetric}

\end{figure*}

\paragraph{Dasymetric Mapping} While our method does not require known
aggregated sums at inference time, it is straightforward to use them
if they are given.  Figure~\ref{fig:censusdasymetric} shows the result
of applying dasymetric mapping using our pixel-level population and
housing density predictions as a guide for redistributing the ground-truth aggregated sums. It is
difficult to evaluate the effectiveness of this approach, since high-resolution density estimates are not available.
However, this approach is promising because the learned CNN is trained in an end-to-end
manner and, hence, does not require any special knowledge of the input
imagery or assumptions about the distribution of objects.

\paragraph{Discussion}

One of the key limitations with our evaluation is that the Census data provides an estimate for the state of the country in 2010, but our imagery is from 2015--2017.  This could lead to a variety of errors, for example replacing a residential neighborhood with a commercial district, constructing a new neighborhood on farm land, or changing the housing density by infill construction.  It may be possible to address these problem by making assumptions on the expected frequency of various types of change, but the best solution is likely to wait for the results of the next Census and capture imagery at roughly the same time.

\section{Conclusions}

We proposed an approach for learning to estimate pixel-level density
functions from high-resolution satellite imagery when only
coarse-grained density aggregates are available for training. Our
approach can be used in conjunction with any pixel-level labeling CNN
using standard deep learning libraries.

We showed that this technique works well on a variety of synthetic
datasets and for a large real-world dataset. The main innovation is
incorporating a layer that replicates the regional aggregation process
in the network in a way that enables end-to-end network optimization.
Since this layer is not required to estimate the pixel-level density,
we can remove it at inference time. The end result is a CNN that
estimates the density with significantly higher effective resolution
than the baseline approach.  We also showed that when additional
information is available about the form of the density function, we
can constrain the learning process, using custom activation functions
or activity regularization, to improve our ability to learn with fewer
samples.

One limitation of our current implementation is that it requires a
region to be fully contained in an input image because it computes the
region aggregation in a single forward pass. When regions are large
and the image resolution is high it may not be possible to do this due
to memory constraints.  One way to overcome this would be to partition
the image and compute the aggregation over multiple forward passes. It
would then be straightforward to compute the backward pass separately
for each sub-image.

For future work, we intend to apply this technique to a variety of
different geospatial density mapping problems, explore additional ways
of incorporating priors, and investigate extensions to structured
density estimation.

\section*{Acknowledgements}

This material is based upon work supported by the National Science Foundation under Grant No. IIS-1553116.  In addition, we would like to acknowledge Planet Labs Inc.\ for providing the PlanetScope satellite imagery used for this work.

\bibliographystyle{ACM-Reference-Format}
\bibliography{biblio}

\end{document}